\documentclass[format=sigconf, dvipsnames]{acmart}

\usepackage{booktabs}

\usepackage{multirow}

\usepackage[labelsep=period]{caption}

\usepackage[]{graphicx}

\usepackage[font=small]{subfig}

\usepackage{amsmath}
\usepackage{amsfonts}

\usepackage[T1]{fontenc}

\usepackage{balance}

\usepackage{siunitx}

\usepackage{enumitem}

\usepackage{soul}

\usepackage{xcolor}

\usepackage{listings}

\definecolor{sbase03}{HTML}{002B36}
\definecolor{sbase02}{HTML}{073642}
\definecolor{sbase01}{HTML}{586E75}
\definecolor{sbase00}{HTML}{657B83}
\definecolor{sbase0}{HTML}{839496}
\definecolor{sbase1}{HTML}{93A1A1}
\definecolor{sbase2}{HTML}{EEE8D5}
\definecolor{sbase3}{HTML}{FDF6E3}
\definecolor{syellow}{HTML}{B58900}
\definecolor{sorange}{HTML}{CB4B16}
\definecolor{sred}{HTML}{DC322F}
\definecolor{smagenta}{HTML}{D33682}
\definecolor{sviolet}{HTML}{6C71C4}
\definecolor{sblue}{HTML}{268BD2}
\definecolor{scyan}{HTML}{2AA198}
\definecolor{sgreen}{HTML}{859900}

\usepackage{algorithm}
\usepackage{algpseudocode}

\begin{document}
	
\renewcommand\footnotetextcopyrightpermission[1]{} 
\pagestyle{plain} 
\pagenumbering{gobble} 
	
	
	\title{Deploying Customized Data Representation and Approximate Computing in Machine Learning Applications}

\author{Mahdi Nazemi}
\affiliation{%
	\institution{University of Southern California}
}
\email{mnazemi@usc.edu}

\author{Massoud Pedram}
\affiliation{%
	\institution{University of Southern California}
}
\email{pedram@usc.edu}


	\begin{abstract}
		Major advancements in building general-purpose and customized hardware have been one of the key enablers of versatility and pervasiveness of machine learning models such as deep neural networks. 
		To sustain this ubiquitous deployment of machine learning models and cope with their computational and storage complexity, several solutions such as low-precision representation of model parameters using fixed-point representation and deploying approximate arithmetic operations have been employed. 
		Studying the potency of such solutions in different applications requires integrating them into existing machine learning frameworks for high-level simulations as well as implementing them in hardware to analyze their effects on power/energy dissipation, throughput, and chip area. 
		Lop is a library for design space exploration that bridges the gap between machine learning and efficient hardware realization. 
		It comprises a Python module, which can be integrated with some of the existing machine learning frameworks and implements various customizable data representations including fixed-point and floating-point as well as approximate arithmetic operations.
		Furthermore, it includes a highly-parameterized Scala module, which allows synthesizing hardware based on the said data representations and arithmetic operations. 
		Lop allows researchers and designers to quickly compare quality of their models using various data representations and arithmetic operations in Python and contrast the hardware cost of viable representations by synthesizing them on their target platforms (e.g., FPGA or ASIC). 
		To the best of our knowledge, Lop is the first library that allows both software simulation and hardware realization using customized data representations and approximate computing techniques. 
		%
		%
		%
	\end{abstract}

	\maketitle
	
 	\section{Introduction}
	Advancements in developing high-performance hardware platforms like GPUs have been a significant enabler for shifting machine learning (ML) models, such as neural networks, from rather theoretical concepts to practical solutions to a wide variety of problems. 
	%
	%
	However, computational and storage complexity of these models has forced the majority of computations to be performed on high-end servers or on the cloud. 
	Meanwhile, the inherent tolerance of many machine learning models to error and approximation has allowed researchers to design systems that benefit from low-precision and approximate computing. 
	This not only reduces computation and storage cost on existing systems, but also enables efficient deployment of such models on resource-constrained platforms such as smartphones and embedded systems. 

	While the majority of machine learning frameworks are widely used in Python,  low-precision and approximate computing techniques are typically implemented in Verilog, VHDL, or C. 
	This has prevented these techniques from being deployed extensively, especially in large-scale models such as deep neural networks (DNNs). 
	Furthermore, training and deployment of complicated machine learning models using hardware simulation tools is extremely burdensome if not impossible. 
	This necessitates redefining approximate computing techniques at a higher level of abstraction for integration with machine learning frameworks. 
	Besides, some of the high-level ideas for reducing storage and computational complexity of deep learning models may not be as effective when implemented in hardware. 
	This motivates introducing new flows for mapping high-level ideas into synthesizable hardware for understanding their actual impact on power, throughput, and area. 

	Lop is a library that bridges the gap between machine learning and efficient hardware realization. 
	It allows users to choose from a variety of data representations and customize the number of bits used for representing model parameters, intermediate values, etc. 
	Furthermore, it allows them to choose how arithmetic operations should be performed on each variable, i.e. the standard for a specific data representation or an approximate implementation available in the library. 
	In other words, Lop introduces a new set of tunable hyperparameters in addition to what machine learning libraries provide, e.g. number of layers, types of layers, number of units per layer, and so on. 
	%
	%
	Lop can be used to answer questions such as the following: 
	\begin{itemize}[nolistsep, leftmargin=*, topsep=0pt]
		\item{Using fixed-point representation, how many bits are required to represent weights, biases, and activations in a deep neural network to reach a target prediction or classification accuracy \cite{courbariaux2016binarized, li2016ternary, venkatesh2017accelerating}? How many bits should be used to represent integral and fractional parts? Can a floating-point representation with fewer number of bits reach the same level of accuracy? How many bits should be used to represent exponent and mantissa? How does fixed-point representation compare with floating-point representation in terms of power/energy consumption, throughput, and area?}
		\item{In a deep neural network, can a layer with lower range of activation values use fewer bits compared to a layer with a higher range of activation values \cite{judd2015reduced}? If so, how many bits are required to represent activations at each layer?}
		\item{During training of a deep neural network, will representing weights and biases with low bit-width during forward pass and high bit-width during backward pass affect the quality of model \cite{zhou2016dorefa, chen2017fxpnet}?}
		\item{How would converting some pre-trained floating-point weights to fixed-point numbers with a predefined bit-width affect prediction accuracy in a deep neural network? Would retraining using the new representation improve the accuracy loss due to conversion?}
		\item{How would replacing some/all multipliers with an approximate multiplier affect prediction or classification accuracy? How much power/energy will be saved by using the approximate multiplier?}
	\end{itemize}	
	
	The remainder of this paper is organized as follows.
	Section~\ref{sec:related-work} reviews some of the related work in low-precision computing in deep neural networks and approximate arithmetic.
	Next, Section~\ref{sec:preliminaries} explains some preliminaries and Section~\ref{sec:framework} details the framework. 
	After that,  Section~\ref{sec:results} presents experimental results and finally, Section~\ref{sec:conclusion} concludes the paper.

	\section{Related Work} \label{sec:related-work}
	
	Designing machine learning models that are efficient in terms of power/energy consumption, area, and/or memory has been widely studied in the past few years \cite{nazemi2017high,lin2018fft,nazemi2018hardware}. 
	Having such efficient models is particularly important for energy- and/or thermal-constrained devices such as smartphones \cite{dousti2015thermtap}. 
	Among different methods for efficient implementation of machine learning models, there has been a considerable amount of work on using various representations such as fixed-point or low bit-width numbers for deep learning applications. 
	This includes using low bit-width floating-point representation for weights and fixed-point representation for activations \cite{lai2017deep}, dynamic fixed-point representation \cite{na2016speeding,gupta2015deep}, fixed-point quantization of clustered weights \cite{mellempudi2017mixed}, binary or ternary weights and activations \cite{courbariaux2016binarized,chen2017fxpnet,li2016ternary,venkatesh2017accelerating,zhou2016dorefa}, and logarithmic representation \cite{miyashita2016convolutional}, to name but a few. 

	Looking at this problem from a hardware design point of view, many researchers have proposed approximate computing techniques for machine learning applications or general-purpose approximate arithmetic units like multipliers and adders that can potentially be used in machine learning applications. 
	Gysel \textit{et~al.} \cite{gysel2016hardware} present a framework that can condense a convolutional neural network by using fixed-point representation for weights and activations. 
	Zhang \textit{et~al.} \cite{zhang2015approxann} introduce a framework that can approximate computation and memory accesses in an artificial neural network by characterizing the impact of different neurons on output quality. 
	Shafique \textit{et~al.} \cite{shafique2016cross} introduce an open-source library of accurate and approximate arithmetic modules and accelerators, however, their modules are implemented in C and VHDL, which prevents their seamless integration into existing machine learning frameworks. 

	Despite the fact that a lot of work has been done in this area, there are no libraries that allow design space exploration using various customizable data representations and approximate arithmetic operations. 
	Additionally, most of the prior work only compare different representations in terms of storage requirement for saving weights, not power/energy efficiency, throughput, and area of their hardware realizations. 
	Lop addresses these issues by integrating various customizable data representations and arithmetic operations into some of the machine learning frameworks for high-level simulations and by providing parameterized hardware implementation of the same data representations and arithmetic operations to compare designs based on power/energy dissipation, throughput, and area figures. 

	\section{Preliminaries} \label{sec:preliminaries}
	In a typical DNN, there are neurons and activation connections (activations for short) among nodes. 
	A neuron is simply a node in the underlying graph of the DNN whereas a synaptic connection is an edge in that graph. 
	Activations carry the output value of a neuron after the dot product calculation, and application of the nonlinear activation function. 
	A neuron also receives hidden variable inputs as weights and biases. 
	These weights and biases are assigned fixed values after the DNN training is completed and remain fixed during the inference. 
	On the other hand, activations carry  a range of values [min, max] over time as a function of the  data that is presented to the inputs of the DNN. 
	Therefore, each weight, bias, or activation in a trained DNN has either a fixed scalar value assignment or a value range assignment. 
	To avoid using too many different data representations (which can result in  a very high implementation cost due to the need to convert back and forth among these representations as we do a forward propagation of input values of the DNN through the network in order to get the classification or recognition result at the output), one should preferably partition the set of nodes, connections, weights, and biases into a small number of different domains where within each domain the choice of data representation and exact vs. approximate arithmetic operation is fixed. 
	In this case, since we need to convert data representation only when we move data across different parts and since the number of parts is small, the said implementation cost overhead can be managed. 
	During training of a DNN, weights and biases will also carry a value range and there are gradients that contain updates to model parameters during backward propagation. 
	We will refer to the set of weights, biases, activations, and gradients as the {\em WBAG set} from here on.

	\section{Proposed Framework} \label{sec:framework}
\begin{figure}[b]
		\centering
		\includegraphics[width=1\columnwidth]{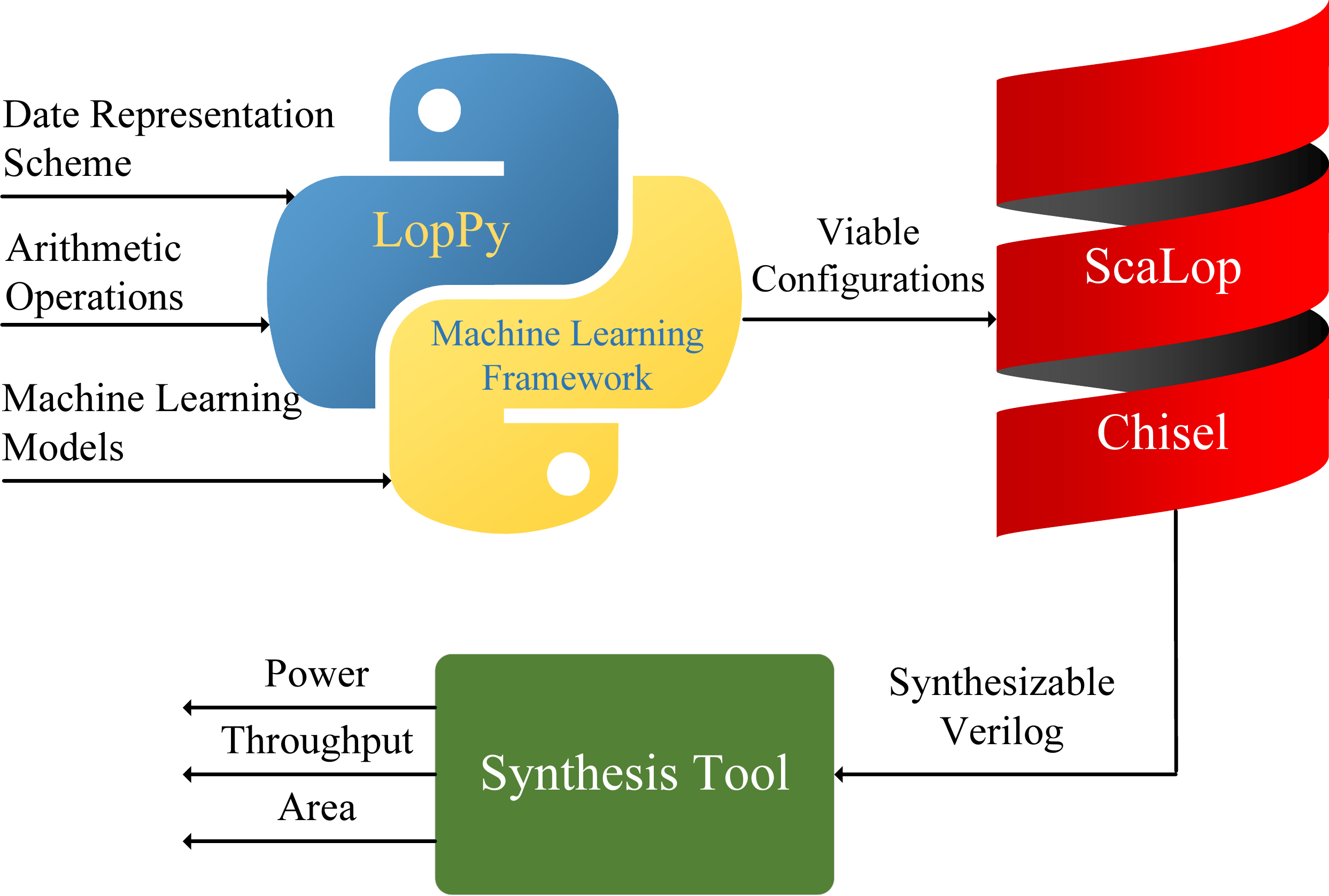}
		\caption{High-level diagram demonstrating Lop in practice. Various models (e.g. networks with different architectures) are provided to machine learning library,  candidate data representations and arithmetic operations are fed to LopPy, viable configurations for each machine learning model are found and fed to ScaLop, synthesizable Verilog files are generated by Chisel, and different metrics are produced by synthesis tool for comparing hardware cost.}
		\label{fig:framework}	
	\end{figure}

	Fig.~\ref{fig:framework} demonstrates a high-level diagram of Lop and its interaction with other libraries and tools. 
	The Python module of Lop, called LopPy, implements various data representations, including fixed-point and floating-point numbers as well as different low-precision and approximate computing methods, including some of the state-of-the-art approximate adders, multipliers, and dividers. 
	LopPy allows these data representations and approximate computing methods to be integrated into some of the existing machine learning frameworks in order to study quality of various ML models under these customized computations. 
	%
	%
	ScaLop is LopPy's counterpart, which is implemented in Scala and interacts closely with Chisel \cite{bachrach2012chisel}. 
	It includes implementation of all data representations and approximate computing methods available in LopPy, which can be used to synthesize customized computations on target platforms such as FPGAs or ASICs. 
	The use of Python and Scala interfaces in Lop enables a high degree of reconfigurability, platform-independence, and programmability. 

\begin{figure}[tb]
		\centering
		\includegraphics[width=0.9\columnwidth]{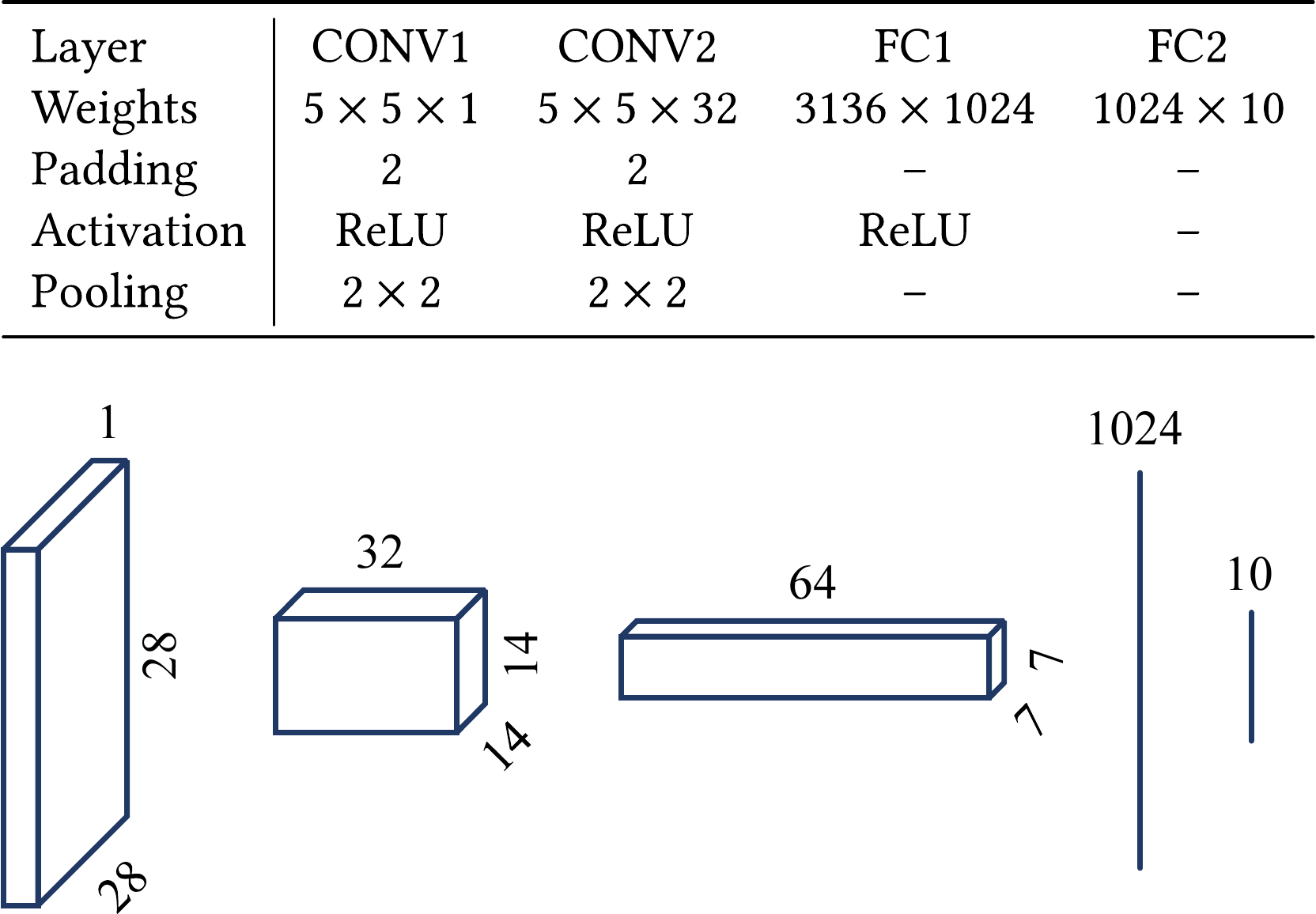}
		\caption{DCNN architecture and shape of activations.}
		\label{fig:dcnn-arch}	
	\end{figure}

	To illustrate different features of Lop throughout the paper, we train a deep convolutional neural network (DCNN) using single-precision floating-point values and use different data representations and arithmetic operations to find an efficient inference engine. 
	Fig.~\ref{fig:dcnn-arch} details the architecture of this DCNN, along with the shape of different layers. 
	The objective of this network is to classify handwritten digits of the MNIST dataset \cite{lecun2010mnist} into one of ten classes. 
	%
	%
	
	\subsection{Data Representation \& Arithmetic Operations}
	This section describes details of different data representations and approximate arithmetic operations implemented in Lop. 
	Various choices of data representations and arithmetic operations may be used with different granularity levels in a machine learning model. 
	Indeed one may use a custom data representation and a corresponding approximate multiplier for the whole DNN or alternatively, use different data representations and mixtures of exact and approximate computing blocks for different parts of the DNN. As an example of the latter scheme, an 8-bit floating-point representation and exact multipliers are employed in the first layer of a DNN, a 12-bit floating-point and truncation-based approximate multipliers are utilized in the second layer, and so on. 
	Similarly, one may use 8-bit fixed-point representation during forward pass and 16-bit fixed-point representation during backward pass of training. 

	\subsubsection{Fixed-Point and Integer Data Representations} 
	Fixed-point representation breaks down bits into an integral part and a fractional part. 
	%
	%
	%
	A fixed-point number can be thought of as an integer number that is multiplied by a scaling factor. 
	Therefore, basic operations on fixed-point numbers like addition and multiplication are similar to integer operations, but take scaling factor into account. 
	This is the main reason that fixed-point arithmetic operations are very efficient in terms of hardware implementation. 
	Integer representation is a special case of fixed-point where the number of fractional bits is set to zero. 

	\subsubsection{Floating-Point Data Representation}
	Floating-point representation uses an exponent and a mantissa to define a number according to  
	\begin{equation*}
		n = mantissa * base^{exponent}.
	\end{equation*}
	This allows floating-point numbers to have a large dynamic range, but introduces complications for hardware implementation. 

	\subsubsection{Arithmetic Operations}
	On top of standard operations for each data representation, Lop implements approximate arithmetic operations inspired by some of the existing methods such as the ones introduced in \cite{hashemi2015drum, imani2017cfpu, narayanamoorthy2015energy, chang2010low}. 
	These operations can be combined with one of the representations described earlier, assuming that the approximate computing method is compatible with the representation. 
	In cases where the work in literature is limited to a specific bit-width, we have generalized the reported work to account for arbitrary bit-widths. 

	\subsection{Exploration Strategy}
	%
	%
	%
	%
	
	The tunable hyperparameters introduced in Lop include choice of data representation, number of bits allocated to each field of data representation, and choice of arithmetic operations.  
	Among these hyperparameters, number of bits allocated to the field that determines value range of WBAG in the DNN under study (i.e. integral part in fixed-point and exponent in floating-point representation) can easily be determined based on network simulations.  
	In practice, the range is usually small, which means only a few bits of the data representation are needed to precisely capture the range of said values. 

	Because customized data representations and approximate arithmetic methods may be used with different granularity levels, values in a network are partitioned into parts where data representation and arithmetic operations within each part are the same. 
	For example, if a network is being optimized layer-wise, each part will include the WBAG set of exactly one layer. Evidently, when there is no need to use different representations and arithmetic operations within two adjacent layers, they can be combined into the same part. 
	Note that when doing DNN training, each element of the WBAG set assumes a range of values that must be determined by doing value dumps of this element as the network is being trained. 
	On the other hand, the weight and bias elements in the WBAG set assume predetermined and fixed values during the inference and only the activations exhibit a non-scalar value range, which is itself determined by dumping activation values for the complete set of training data (note that gradients do not matter and are ignored during the inference; so in fact we are only interested in the WBA set). 
	Table~\ref{table:range} summarizes value ranges for the network of Fig.~\ref{fig:dcnn-arch} assuming layer-wise optimization. 

	\begin{table}[h]
		\centering
		\captionsetup{justification=centering}
		\caption{Value range of weights, biases, and activations in each layer of the network of Fig.~\ref{fig:dcnn-arch}}
		\label{table:range}
		\resizebox{\columnwidth}{!}{%
			\begin{tabular}{l cccc}
				Layer            & CONV1                     & CONV2                     & FC1                   & FC2                   \\
				\midrule
				Range            & [-1.45, 1.15]             & [-3.33, 2.45]             & [-9.85, 6.80]         & [-28.78, 35.76]
			\end{tabular}
		}
	\end{table}

	Given the value range of each part, one can calculate the number of bits that are required for representing that range. 
	For example, to support the value range of the first fully-connected layer (FC1), a fixed-point representation requires four bits in the integral part (in a sign-magnitude format). 
	However, because the value range of partial sums may be greater than the value ranges mentioned in Table~\ref{table:range}, we extend the number of bits to a larger interval to ensure correct arithmetic operations. 
	As a result, instead of setting the number of bits for representing the integral part of values in FC1 to four, the number of bits will be chosen from an interval that is lower bounded by four, e.g. [4, 7]. 
	We must add to this bit count another bit to represent the sign. 
	A similar analysis may be performed for representing the exponent in a floating-point representation. 

	Unfortunately, the value ranges do not help us determine lower or upper bounds on the number of bits needed for the part of data representation that determines the computational accuracy (i.e. fractional part in the fixed-point and mantissa in the floating-point representation). 
	Therefore, here we resort to enumerating the bit count of this part of the data representation in some predefined interval, e.g., [4, 12]. 
	We refer to the intervals for different parts of the data representation as bit count intervals or BCIs for short. 
	It should be noted that using exact or approximate arithmetic operations affects BCIs. 
	For example, an approximate floating-point multiplier may need a higher number of bits in mantissa to achieve acceptable classification or prediction accuracy. 

	To find the best data representation and arithmetic operation for each part of the partition, the parts in the DNN are sorted topologically, starting from the input layer and moving towards the output layer. 
	After that, the data representation and arithmetic operation for each part is found according to its BCI such that it minimizes hardware cost subject to bounded loss in classification or prediction accuracy. 
	Throughout this process, the parts that come before the part under study are implemented with their optimized data representation and arithmetic operation while the parts that come after the part under study are implemented with full precision and exact operations to ensure they do not introduce additional loss in classification or prediction accuracy. 
	This process continues till all parts are optimized. 
	Optionally, a second pass of optimization can be performed for quality recovery. 
	During this pass, the objective is to maximize classification or prediction accuracy subject to bounded increase in hardware cost. 
	Throughout this process, the parts are optimized in the same order that they were processed during the first pass of optimization. 
	The difference, though, is that the parts that come after the part under study are implemented with their optimized representation that was found during the first pass of optimization. 
	The bounded increase in hardware cost can be translated to some constraints on BCIs. 
	For example, the representation for each part may only use one additional bit compared to the representation that was found during the first pass of optimization. 
	%
	
%
%
%
%
%

	\subsection{LopPy}
	LopPy implements a Numeric class in Python for each of the data representations described earlier. 
	These representations can be customized by user in terms of number of bits that are allocated to a specific representation, e.g. number of bits to represent exponent and mantissa in a floating-point number. 
	The implementation also includes arithmetic operations such as multiplication/division, addition/subtraction, exponentiation, comparison, etc. and is compatible with both Python 2 and Python 3. 
	Additionally, there are behavioral implementations of approximate arithmetic modules that can be combined with one of the data representations. 
	For example, a user may choose an 8-bit floating-point representation that allocates one bit to sign, four bits to exponent, and three bits to mantissa. 
	Furthermore, he/she may replace the standard multiplication and division with an approximate method that is compatible with floating-point representation, but keep other arithmetic operations untouched. 
	%
	
	%
	All implemented Numeric classes accept strings, floating-point numbers, and integers in their constructors. 
	This enables using different configurations for representing numbers across layers of a deep neural network, during forward and backward passes of training, etc. 
	For example, when two related layers of a deep neural network are not compatible in terms of data representation or number of bits per field, values in the input layer will be converted to an intermediate floating-point representation and later, to a representation that is compatible with the output layer. 
	%
	%

	The following code snippet demonstrates an example of inference on the network of Fig.~\ref{fig:dcnn-arch} with 12-bit and 16-bit fixed-point values for convolutional and fully-connected layers, respectively. 
	For each representation, the number of integral and fractional bits is set separately. 
\begin{lstlisting}[language=Python]
from loppy.fixedpoint import *
from loppy.utils import *

# "network" is the trained network of Fig. 2
dtype = FixedPoint
context = {'int_width' : 4, 'frac_width' : 8}
for layer in network.layers[0:2]: # CONV layers
	layer.params = to_dtype(layer.params, dtype, context)

context = {'int_width' : 6, 'frac_width' : 10}
for layer in network.layers[2:4]: # FC layers
	layer.params = to_dtype(layer.params, dtype, context)
	
# forward propagation using new data representations
pred = network.forward(inputs)
\end{lstlisting}

	There are a few optimizations in LopPy that increase performance and reduce memory usage. 
	The first one is a result of LopPy's compatibility with Cython. 
	Cython produces a standard Python module that can be imported in other modules. 
	However, the original Cython-compatible Python module is translated into C, which is further compiled to machine code, resulting in faster code. 
	Cython programs usually consume fewer computing resources such as processing cycles and memory. 
	On our benchmarks, we achieved 2x performance improvement by using Cython-generated modules of our Numeric classes. 
	As a result, a user may use the Python variant of LopPy for quick and easy development and testing and use the Cython variant in production code. 

	Another performance advantage comes from the use of \texttt{\_\_slots\_\_} for defining instance variables. 
	This restricts the valid set of attributes to the ones listed in \texttt{\_\_slots\_\_} and therefore, allows efficient storage of attributes in an array. 
	It has been shown that using \texttt{\_\_slots\_\_} can increase performance by 15-30\%  \cite{slots-stackoverflow}. 
	Additionally, using \texttt{\_\_slots\_\_} leads to a low, predictable memory usage, which is in contrast to using a \texttt{\_\_dict\_\_} that is the default way of storing instance variables in Python. 
	It has been shown that using \texttt{\_\_slots\_\_} can improve memory footprint by around 70\% compared to using \texttt{\_\_dict\_\_} \cite{slots-stackoverflow-2}. 

	It is worth mentioning that application of LopPy goes beyond machine learning. 
	For example, a user that has used SciPy to solve a problem in signal processing, image processing, or linear algebra may use LopPy to see how an objective is affected when a different data representation or an approximate computing technique is applied. 

	\subsection{ScaLop} \label{sec:scalop}
	ScaLop has a similar implementation to LopPy, but is used for hardware design and analysis. 
	It defines the same data representations and approximate computing methods in such a way that is compatible with Chisel. 
	While the majority of prior work compare various data representations in terms of memory requirement for storing weights, ScaLop allows full comparison of various configurations in terms of power consumption, throughput, and area due to its seamless integration into existing systems implemented in Chisel. 
	One of the advantages of Chisel that makes it suitable for our framework is its automatic width inference. 
	Automatic width inference allows users to modify bit-width of data representation without needing to manually modify other dependent modules. 
	Furthermore, FIRRTL, the intermediate representation that is generated during RTL generation, can introduce a great degree of compile-time reconfigurability. 
	Additionally, Chisel can generate both synthesizable Verilog files for synthesis on target platforms and C++ representation of circuits for fast simulations using Verilator \cite{snyder2004verilator}. 

	The following code snippet illustrates an example of defining a processing element (PE) that consists of a multiplier and an adder in which inputs and outputs are fixed-point numbers with six bits in integral part and eight bits in fractional part. 
	These arithmetic operations may be replaced with another data representation or approximate arithmetic unit that is available in the library. 
\begin{lstlisting}[language=Scala]
import chisel3._
import chisel3.util._
import scalop.{FixedMul, FixedAdd}

class PE(width : Int) extends Module {
  /* define primary inputs/outputs here */

  // Multiplier
  val mul = Module(new FixedMul(6, 8)).io
  /* connect input/output ports */


  // Adder
  val add = Module(new FixedAdd(6, 8)).io
  /* connect input/output ports */

  /* assign output(s) */
}
\end{lstlisting}
	
	It should be noted that one may integrate ScaLop modules into an existing Verilog design without having the design implemented in Chisel. 
	Verilog files for standard and approximate operations can be generated using Chisel and replaced with corresponding modules in Verilog design. 
	As a result, ScaLop may be used directly in existing Chisel projects or indirectly, through generation of Verilog modules, into existing Verilog designs. 

	It is worth mentioning that Chisel is a high-level language that may not create the most efficient Verilog implementation. 
	As a result, the estimated hardware cost is an upper bound and the user may need to fine tune the Verilog code to achieve higher power/energy efficiency, increased throughput, and lower area. 

	\subsection{Extending Lop}
	Lop allows users to easily define new data representations and arithmetic operations. 
	For example, suppose that a user wants to implement a neural network where weights and activations are 0/1 binary values and multiply operations are replaced with XNOR, e.g. a network similar to \cite{courbariaux2016binarized}. 
	Because existing libraries implement operations such as two-dimensional convolutions using multiply and add operations, the user needs to define convolutions from scratch to use XNOR instead of multiplication. 
	This includes transforming inputs into a Teoplitz matrix, implementing different strides and paddings, etc. 

	Lop provides a simple solution to this problem where the user can define a new data representation based on fixed-point representation in which the number of integral bits is one and there are no fractional bits, hence achieving binary values. 
	Furthermore, the multiply operation is overridden to implement XNOR instead of multiplication. 
	As a result, when a machine learning library applies a multiplication within convolution operation, XNOR is called under the hood. 
	Therefore, the user may use functionalities of the machine learning library without redefining basic operations. 
	The following code snippet illustrates one such implementation. 
\begin{lstlisting}[language=Python]
from loppy.fixedpoint import *

class BinXNOR(FixedPoint):
	# create a new object and initialize it here
	
	def __mul__(self, other):
		# create "res" object with the same context as self
		res._num = self._num ^ other._num # XOR
		res.__invert__() # NOT
		return res
	
	# Python's built-in NOT calculates 1's complement
	def __invert__(self):
		self._num = 1 - self._num
\end{lstlisting}
	
	\section{Experimental Results} \label{sec:results}
	
	\subsection{Software Simulation} \label{sec:sw-results}
	The trained model of Fig.~\ref{fig:dcnn-arch} is able to classify test data with 99.1\% accuracy using single-precision floating-point values. 
	This classification accuracy is considered as baseline and all other accuracies are normalized to this value for easier comparison. 
	This section explains how other data representations and approximate arithmetic operations may be used to design an efficient inference engine for this network. 
	Table~\ref{table:dtypes} summarizes data representations and approximate computing methods used in this section. 

    \begin{table}[h]
	\centering
	\captionsetup{justification=centering}
	\caption{Summary of notation}
	\label{table:dtypes}
	\resizebox{\columnwidth}{!}{%
		\begin{tabular}{l | l}
			Notation            & Description  \\
			\hline
			FL(e,~m)            & Floating-point representation with $e$ exponent bits and $m$ mantissa bits.                 \\
			I(e,~m)               & Similar to FL(e,~m), but with approximate multiplier based on \cite{imani2017cfpu}.       \\
			FI(i,~f)                & Fixed-point representation with $i$ integral bits, and $f$ fractional bits.                             \\
			H(i,~f,~t)           & Similar to FI(i,~f), but with approximate multiplier of width $t$ based on \cite{hashemi2015drum}.
		\end{tabular}
	}
	\end{table}
	
	Table~\ref{table:sw-float} summarizes normalized classification accuracy for some of the explored customized computations based on floating-point representation. 
	It includes representations with different bit-widths in each layer of the network and other configurations where some or all multiply operations are replaced with approximate multipliers. 
	Those customized computations that achieve the same classification accuracy as baseline, i.e. 100\% relative accuracy, are selected for hardware realization in the next step. 

	\begin{table}[h]
		\centering
		\captionsetup{justification=centering}
		\caption{Classification accuracy for different customized computations based on floating-point representation}
		\label{table:sw-float}
		\resizebox{\columnwidth}{!}{%
			\begin{tabular}{ccccc}
				\toprule
			    \multicolumn{4}{c}{Layers}										  & \multirow{2}{*}{Relative Accuracy} \\
				\cmidrule{1-4}
				CONV1        & CONV2        & FC1          & FC2           & {}               \\
				\midrule
				FL(4, 8)      & FL(4, 9)        & FL(4, 8)  & FL(4, 9)   & 98.98\%    \\
				\textbf{FL(4, 9)}      & \textbf{FL(4, 9)}        & \textbf{FL(4, 9)}  & \textbf{FL(4, 9)}   & \textbf{100\%}    \\
				I(4, 8)          & I(4, 9)           & I(4, 8)     & I(4, 9)       & 94.90\%    \\
				I(4, 9)         & I(4, 9)            & I(4, 9)     & I(4, 9)       & 94.90\%    \\
				\textbf{I(5, 10)}        & \textbf{I(5, 10)}          & \textbf{I(5, 10)}     & \textbf{I(5, 10)}       & \textbf{100\%}    \\
				\bottomrule
			\end{tabular}
		}
	\end{table}

	Similarly, Table~\ref{table:sw-fixed} summarizes normalized classification accuracy for some of the explored customized computations based on fixed-point representation. 
	Among customized computations that meet baseline classification accuracy, FI(6, 8) has the lowest number of bits and does not have complications such as leading-one detector and barrel shifter that is used in \cite{hashemi2015drum}. 
	As a result, this data representation is selected for hardware realization in the next step. 

	\begin{table}[h]
		\centering
		\captionsetup{justification=centering}
		\caption{Classification accuracy for different customized computations based on fixed-point representation}
		\label{table:sw-fixed}
		\resizebox{\columnwidth}{!}{%
			\begin{tabular}{ccccc}
				\toprule
				\multicolumn{4}{c}{Layers}										  & \multirow{2}{*}{Relative Accuracy} \\
				\cmidrule{1-4}
				CONV1        & CONV2        & FC1          & FC2           & {}               \\
				\midrule
				FI(5, 8)        & FI(5, 8)        & FI(6, 8)  & FI(6, 8)   & 98.98\%    \\
				FI(6, 8)        & FI(6, 8)        & H(8, 8, 14)  & H(8, 8, 14)   & 100\%    \\
				H(6, 8, 12)        & H(6, 8, 12)        & H(8, 8, 14)  & H(8, 8, 14)   & 100\%    \\
				\textbf{FI(6, 8)}        & \textbf{FI(6, 8)}        & \textbf{FI(6, 8)}  & \textbf{FI(6, 8)}   & \textbf{100\%}    \\
				\bottomrule
			\end{tabular}
		}
	\end{table}

	\subsection{Hardware Realization}
	To compare hardware cost of various implementations using different data representations and arithmetic operations, we take a similar approach to \cite{sharma2016dnnweaver} where the deep neural network is mapped to a set of processing elements (i.e. the datapath) and proper control signals are generated to schedule computations on PEs. 
	The difference, though, is that our implementation consists of 500 PEs where the multiplier and adder inside the PE operate on customized data representations and may be exact or approximate. 
	The target FPGA is part of Arria 10 family, which includes 427,200 adaptive logic modules (ALMs), 55,562,240 bits of block RAM, and 1518 DSP blocks. 

	Table~\ref{table:fpga-results} compares hardware cost of implementing this datapath using different data representations and approximate arithmetic operations that were found viable in the previous section (Tables~\ref{table:sw-float} ~and~\ref{table:sw-fixed}). 
	The representation shown in the table is used for all layers of the network. 
	The table also includes two baseline implementations using single-precision and half-precision floating-point representations, respectively (float32 and float16). 

    \begin{table}[h]
	\centering
	\caption{Hardware Cost of Various Implementations}
	\label{table:fpga-results}
	\resizebox{\columnwidth}{!}{%
		\begin{tabular}{lccccc}
			\toprule
			\multirow{2}{*}{Representation}          & ALMs                           & DSPs                            & Clock       & Power     & Energy Efficiency \\
			{}                                                                           & count (util. factor)  &  count (util. factor) &  (MHz)     & (W)           & (Gops/J) \\
			\midrule
			float32                                                               & 209,805 (49\%)     & 500 (33\%)              & 94.41       & 12.38      & 3.81 \\
			float16                                                                & 101,644 (24\%)      & 500 (33\%)              & 113.86     & 7.30        & 7.80 \\
			FL(4, 9)                                                             & 93,500 (22\%)        & 500 (33\%)              & 115.89     & 6.68        & 8.67 \\
			I(5, 10)                                                               & 92,111 (22\%)          & 0 (0\%)                       & 116.80     & 6.28        & 9.30 \\
			FI(6, 8)                                                              & 15,452 (4\%)            & 500 (33\%)              & 201.13     & 4.90        & 20.52 \\
			\bottomrule
		\end{tabular}
	}
\end{table}
	
	There are a few interesting conclusions that can be made from Table~\ref{table:fpga-results}. 
	First of all, it shows the potential of Lop in integrating approximate computing techniques into large-scale systems. 
	The I(5,~10)-based realization of the said datapath achieves the baseline accuracy without using any DSP blocks from the FPGA and consumes 50\% and 14\% less power compared to single-precision and half-precision floating-point baselines, respectively. 
	While reference \cite{imani2017cfpu} shows the effectiveness of this approximate computing method in smaller benchmarks, Lop allows using it in a convolutional neural network, which has a much more complicated design. 
	Moreover, the I(5,~10)-based realization  has a peak clock speed that is 24\% and 3\% higher than the said baselines, respectively. 
	The reported data shows that the I(5,~10)-based realization achieves an overall energy efficiency (e.g., MIPS/Watt or ops/Joule) increase of 144.09\% and 19.23\% over the baseline implementations, respectively. 
	Second, it can be observed that the fixed-point representation FI(6,~8) consumes about half to one third power compared to baseline implementations, can operate twice as fast, and utilizes considerably fewer ALMs. 
	Additionally, it improves energy efficiency by 438.58\% and 163.08\% compared to the baselines. 
	However, compared to I(5,~10), it requires 500 DSP blocks, which may be considered as a disadvantage. 
	Finally, FL(4,~9) can improve power consumption by 46\% and 9\% compared to single-precision and half-precision floating-point, respectively, while it achieves the same prediction accuracy. 
	Additionally, for this representation, the number of ALMs and clock frequency is slightly better than those of half-precision floating-point representation. 

	This information can be used to decide the appropriate data representation and approximate computing methods based on a platform's available resources. 
	For example, if a platform has a limited number of DSP blocks, then I(5,~10) is a good choice because it is a multiplier-free implementation. 
	On the other hand, if power consumption, throughput, and consumed ALMs are of higher importance, a fixed-point representation is preferred. 
	Finally, if floating-point representation is desired, then FL(4,~9) has the lowest number of bits that can achieve baseline classification accuracy. 

	\balance

	\section{Conclusion} \label{sec:conclusion}
	In this work, we presented Lop, a library for cross-dimensional comparison of deploying different customized data representation and approximate computing techniques. 
	Lop consists of a Python module, called LopPy, that allows design space exploration by using various data representations and approximate computing techniques. 
	The counterpart of LopPy for hardware analysis, called ScaLop, is compatible with Chisel and allows designers to compare the hardware cost of their designs for configurations that are found viable by using LopPy. 
	The use of Python and Scala interfaces in this framework enables a high degree of reconfigurability, platform-independence and programmability. 
	While Lop is mainly targeted at machine learning applications, it can be used in a wide variety of applications that involve low-precision and approximate computing such as near-sensor computing. 
	%

	
	\section*{Acknowledgements}
	This research was sponsored in part by a contract from the National Science Foundation.
	\bibliographystyle{unsrt}
	{\footnotesize
		\bibliography{lop}
	}
	
\end{document}